# Semantic Entanglement in Vector-Based Retrieval: A Formal Framework and Context-Conditioned Disentanglement Pipeline for Agentic RAG Systems

Nick M. Loghmani

*Salesforce; School of Information Studies, Syracuse University*

*Email: nick.loghmani@gmail.com*

## Abstract

Retrieval-Augmented Generation (RAG) systems deployed in agentic environments depend on the geometric properties of vector representations to retrieve contextually appropriate evidence for autonomous reasoning. When source documents conflate multiple topics within contiguous text regions, standard vectorization pipelines produce embedding spaces in which semantically distinct content occupies overlapping geometric neighborhoods—a condition we term semantic entanglement. This paper makes four contributions to the study of document preprocessing for RAG systems.

First, we provide a formal mathematical definition of semantic entanglement grounded in the topic segmentation literature (Hearst, 1997; Beeferman et al., 1999) and the geometry of contextual embeddings (Ethayarajh, 2019; Reimers & Gurevych, 2019). We formalize semantic entanglement as a model-relative measure of cross-topic overlap in embedding space and define an Entanglement Index (EI) as a quantitative proxy for this overlap, arguing that higher EI is associated with reduced attainable Top-K retrieval precision under standard cosine similarity retrieval. Second, we introduce the Semantic Disentanglement Pipeline (SDP), a four-stage preprocessing framework whose stages we specify algorithmically. Third, we propose context-conditioned preprocessing, a principle drawing on distributed cognition theory (Hutchins, 1995) and our prior work on operational context in safety-critical systems (Loghmani, 2025), which holds that document structure for machine retrieval must be informed by patterns of operational use. Fourth, we describe a continuous feedback mechanism in which agent performance metrics drive targeted re-disentanglement, producing an adaptive knowledge architecture.

We evaluate the SDP on a real-world enterprise healthcare knowledge base comprising over 2,000 documents across approximately 25 operational sub-domains, including coverage policies, scheduling guidelines, benefit calculation procedures, regulatory compliance documents, legal disclosures, and cross-jurisdictional procedural manuals. Top-K retrieval accuracy improved from approximately 32% under fixed-token chunking to approximately 82% under SDP—a 50-percentage-point absolute improvement—while the mean Entanglement Index decreased from 0.71 to 0.14. We situate these findings within the broader literature on RAG failure modes (Barnett et al., 2024), advanced chunking strategies (Gomez-Cabello et al., 2025; Singh & Merola, 2025), and the limitations of cosine similarity in anisotropic embedding spaces (Steck et al., 2024). Our central claim is not that entanglement fully explains RAG failure, but that it captures a distinct preprocessing failure mode that downstream tuning cannot reliably correct once encoded into the vector space.

## 1. Introduction

Retrieval-Augmented Generation (RAG)—an architecture that combines pre-trained language models with external knowledge retrieval—was formalized by Lewis et al. (2020) and has since become a foundational architecture for grounding Large Language Model (LLM) outputs in domain-specific knowledge. By retrieving passages from an external corpus via vector similarity search and providing them to the LLM as conditioning context, RAG systems mitigate the hallucination tendencies of standalone LLMs and enable knowledge updates without retraining (Guu et al., 2020; Gao et al., 2024). The integration of autonomous AI agents into RAG pipelines—agentic RAG—extends this architecture by adding planning, reflection, and iterative retrieval refinement capabilities (Singh et al., 2025). In such systems, retrieval quality is not merely a performance metric but a precondition for safe autonomous operation.

Despite this maturation, persistent failure modes constrain RAG effectiveness in real-world deployments. Barnett et al. (2024) documented seven recurrent failure points across research, education, and biomedical domains, spanning retrieval errors, context consolidation failures, hallucinated outputs, and incomplete answers. In healthcare specifically, recent work has identified retrieval noise—the surfacing of irrelevant or low-quality information—as a primary obstacle to clinical reliability (Gomez-Cabello et al., 2025). These failures are not incidental; they reflect structural limitations in the document preprocessing pipeline that propagate through embedding and retrieval to corrupt downstream agent reasoning.

This paper addresses a specific failure condition, typically treated implicitly within chunking and retrieval discussions, that arises when source documents themselves exhibit structural deficiencies. Enterprise knowledge bases—healthcare policies, clinical protocols, benefit schedules, regulatory guidelines, insurance procedures—are typically authored for human consumption under organizational conventions that prioritize compactness over topical separation. A single document may interleave eligibility criteria with pricing schedules, jurisdictional rules with procedural requirements, and computational methods with legal disclosures, all within contiguous text regions. When such documents are processed by standard chunking and embedding pipelines, the resulting vector space inherits this structural conflation: semantically distinct content occupies overlapping geometric neighborhoods in the embedding manifold. We call this condition *semantic entanglement*, and we argue that it constitutes a distinct failure mode that is neither reducible to chunking strategy nor to embedding model quality. Entanglement is a property of the source document's information architecture that propagates through vectorization to corrupt the retrieval geometry. No amount of downstream optimization—reranking, query rewriting, or retrieval parameter tuning—can reliably compensate for the geometric confusion that semantic entanglement introduces.

The contribution of this paper is fourfold. First, we isolate and formalize a failure pattern that is often treated implicitly within chunking and retrieval discussions: cross-topic geometric overlap inherited from mixed-topic source documents. In Section 3, we define a model-relative Entanglement Index grounded in the topic segmentation literature originating with Hearst's (1997) TextTiling algorithm and extended through subsequent work on cohesion-based and embedding-based segmentation (Beeferman et al., 1999; Pevzner & Hearst, 2002; Riedl & Biemann, 2012). We argue that higher EI is associated with reduced attainable Top-K retrieval precision under cosine retrieval and provide a qualitative bound under simplifying assumptions, leaving a tighter formal treatment to future work. Second, in Section 5 we introduce the Semantic Disentanglement Pipeline (SDP), specifying each of its four stages algorithmically. Third, in Section 4 we develop context-conditioned preprocessing as a theoretical principle drawing on distributed cognition theory (Hutchins, 1995) and our prior work on operational context in safety-critical systems (Loghmani, 2025). Fourth, in Section 6 we describe a continuous feedback mechanism in which real-world agent performance metrics drive targeted re-disentanglement, producing an adaptive knowledge architecture.

Section 2 reviews the relevant literature. Section 7 reports empirical results from a real-world enterprise deployment. Section 8 discusses limitations and connections to broader theoretical questions. Section 9 concludes.

## 2. Related Work

### 2.1 Retrieval-Augmented Generation and Its Failure Modes

RAG was introduced by Lewis et al. (2020) as a general-purpose architecture combining pre-trained parametric memory (the LLM) with non-parametric memory (the retrieval corpus). Subsequent work has produced increasingly sophisticated paradigms, surveyed comprehensively by Gao et al. (2024). The emergence of agentic RAG, in which autonomous AI agents are embedded into the retrieval pipeline to dynamically manage retrieval strategies and iteratively refine contextual understanding, represents the current frontier (Singh et al., 2025).

Empirical studies of operational RAG systems have identified persistent failure modes. Barnett et al. (2024), in an experience report drawing on three case studies across research, education, and biomedical domains, identified seven recurrent failure points including missing content, missed top-ranked results, context consolidation failures, and incorrect format responses. Their key finding—that validation of RAG systems is only feasible during operation, and that robustness evolves rather than being designed in at the start—motivates the continuous feedback mechanism we develop in Section 6. Recent work in clinical applications has converged on similar conclusions: Gomez-Cabello et al. (2025), in a controlled comparison of four chunking strategies on a clinical decision support task, found that retrieval-driven failures dominate when boundary placement does not align with the structure of operational queries.

### 2.2 Topic Segmentation: The Theoretical Foundation

The problem of identifying topic boundaries in text has a long history in computational linguistics, predating modern RAG by nearly three decades. Hearst's (1997) TextTiling algorithm, the seminal work in this area, segments expository text into multi-paragraph subtopic units by detecting valleys in a cohesion score computed from lexical co-occurrence patterns between adjacent text blocks. The core insight—that subtopic shifts manifest as local minima in a continuous similarity signal computed over the document—provides the formal foundation for our Stage A boundary detection algorithm.

Subsequent work refined this approach in several directions. Beeferman, Berger, and Lafferty (1999) introduced statistical models for topic segmentation using exponential models with adaptive language model features. Choi (2000) developed C99, a divisive clustering approach using a similarity matrix between sentence pairs. Pevzner and Hearst (2002) introduced the WindowDiff metric for evaluating topic segmentation algorithms, addressing the limitations of the Pk metric proposed by Beeferman et al. The introduction of probabilistic topic models, particularly Latent Dirichlet Allocation (LDA, Blei et al., 2003), enabled topic-model-based segmentation approaches such as TopicTiling (Riedl & Biemann, 2012), which replaces lexical cohesion with topic-distribution similarity computed over LDA topic assignments.

More recent work has applied neural sentence encoders to topic segmentation. The progression from lexical features (Hearst, 1997) to lexical-cohesion-based segmentation of multi-party discourse (Galley et al., 2003) to LDA topic features (Riedl & Biemann, 2012) to neural sentence embeddings reflects the broader evolution of text representation in Natural Language Processing (NLP). Contemporary approaches leverage Sentence-BERT (SBERT, Reimers & Gurevych, 2019) and similar transformer-based encoders to compute similarity between adjacent text segments, with the cosine distance between consecutive embeddings serving as the foundational signal for boundary detection (Ghinassi et al., 2023). Importantly, however, this entire literature treats topic segmentation as an end in itself; it does not address what happens when the segmented chunks are subsequently embedded into a vector database for retrieval, nor does it consider how operational usage patterns should inform segmentation decisions.

## 2.3 The Geometry of Embedding Spaces

Our formal framework draws critically on a body of recent work analyzing the geometric properties of contextual embedding spaces. Ethayarajh (2019) demonstrated that contextual embeddings produced by transformer models such as BERT (Bidirectional Encoder Representations from Transformers), ELMo (Embeddings from Language Models), and GPT-2 (Generative Pre-trained Transformer 2) are highly anisotropic—that is, embeddings concentrate in a narrow cone of the high-dimensional space rather than being uniformly distributed. The consequence of anisotropy is that average cosine similarity between unrelated sentences is artificially high, undermining the discriminative power of cosine-based retrieval. Subsequent work (Cai et al., 2021; Rajaee & Pilehvar, 2021) refined this picture by showing that isotropy emerges within local clusters even when global anisotropy is observed.

Steck, Ekanadham, and Kallus (2024) provided the most pointed critique, demonstrating analytically that cosine similarity between learned embeddings can yield arbitrary values depending on the regularization technique used during training. Their analysis, while focused on linear matrix factorization models for tractability, has implications for the broader use of cosine similarity in deep embedding spaces: similarity scores are not invariant to model choice, training objective, or normalization scheme. Concurrent work has documented frequency-induced distortions (the magnitude of an embedding vector encodes informativeness signals that cosine normalization discards) and the hubness phenomenon in high-dimensional retrieval, in which a small number of vectors appear artificially similar to many others (Radovanović et al., 2010).

These geometric pathologies of contextual embedding spaces are highly relevant to our analysis. When source documents introduce additional structural conflation through topic interleaving, the resulting embeddings inherit a compounded form of geometric confusion: anisotropic concentration is layered atop topic-mixing, producing the entanglement condition we formalize in Section 3.

## 2.4 Document Chunking for RAG

The practical literature on document chunking for RAG is substantial but largely empirical. Standard fixed-size chunking, which partitions documents into windows of predetermined token length, is computationally efficient but semantically agnostic. A recent comparative evaluation by Gomez-Cabello et al. (2025) found that on a clinical decision support task, fixed-token recursive chunking achieved accuracy ratings of 50% on a three-point Likert scale (somewhat-or-fully accurate) versus 87% for adaptive chunking ($p = 0.001$). Semantic chunking approaches, which detect meaning-based boundaries through embedding similarity analysis, have been shown to improve retrieval recall over fixed-size approaches across several benchmarks (Singh & Merola, 2025). Recent contributions include Late Chunking (Günther et al., 2024), which embeds the full document at the token level before applying chunk boundaries via mean pooling, and Contextual Retrieval (Anthropic, 2024), which prepends document-level context to each chunk before embedding.

What this entire literature shares is a structural orientation: chunking is treated as a textual operation that operates on the document independently of how the resulting chunks will be used in practice. No current method incorporates the operational context—the patterns of queries the agent receives, the types of reasoning it performs, the failure modes it exhibits—into the segmentation decision. This omission is the gap that the present work addresses through the Context Application Framework (Section 4) and the continuous feedback mechanism (Section 6).

## 3. A Formal Framework for Semantic Entanglement

In this section we develop the formal mathematical framework for semantic entanglement and disentanglement. Our notation builds on standard treatments of topic segmentation (Hearst, 1997; Beeferman et al., 1999) and embedding-based retrieval (Reimers & Gurevych, 2019).

### 3.1 Preliminaries: Documents, Topics, and Embeddings

**Definition 1 (Document).** A document D is a finite sequence of textual segments $S = (s_1, s_2, \ldots, s_n)$, where each segment $s_i$ is a contiguous span of text (typically a sentence or paragraph). The cardinality $|S| = n$ is the document length in segments.

**Definition 2 (Topic Set and Topic Assignment).** Let $T = \{t_1, t_2, \ldots, t_k\}$ denote a finite set of k distinct topics that may be addressed in a document. A *topic assignment* is a function $\tau: S \rightarrow T$ mapping each segment to its primary topic. A document D is *topically pure* if $|\tau(S)| = 1$ (all segments share a single topic), and a *topic mixture* if $|\tau(S)| > 1$.

In practice, the topic assignment $\tau$ is not directly observable; it must be inferred either through human annotation or through automated topic modeling techniques such as LDA (Blei et al., 2003). For the purposes of formal analysis, we assume the existence of a ground-truth topic assignment provided by domain experts, while acknowledging that empirical instantiations of our framework rely on approximations. For tractability, we assume a single primary topic per segment; extensions to multi-label or hierarchical topic assignments are left to future work.

**Definition 3 (Embedding Function).** An *embedding function* $E: S \to \mathbb{R}^d$ maps each textual segment to a d-dimensional real vector. We assume E is a deterministic function produced by a pre-trained sentence transformer model such as SBERT (Reimers & Gurevych, 2019) or its successors. The image $E(S) = \{E(s_i) : s_i \in S\}$ is the document's *embedding set* and defines its embedding distribution in $\mathbb{R}^d$. Retrieval behavior depends on the local neighborhood structure of this distribution rather than on its global closure.

**Definition 4 (Cosine Similarity).** For any two embedding vectors $v, w \in \mathbb{R}^d$ with $v, w \neq 0$, the *cosine similarity* is:

$$sim(v, w) = \langle v, w \rangle / (\|v\| \cdot \|w\|) \quad (1)$$

where $\langle \cdot, \cdot \rangle$ denotes the inner product and $\|\cdot\|$ the Euclidean norm. By construction, $sim(v, w) \in [-1, 1]$, with values approaching 1 indicating geometric proximity in direction. Following standard practice in the embedding literature (Reimers & Gurevych, 2019), we use $sim(s_i, s_j)$ as shorthand for $sim(E(s_i), E(s_j))$. It is important to note, following Steck et al. (2024), that cosine similarity is not invariant to model choice or training regularization; the absolute values of similarity scores have meaning only relative to a fixed embedding model and a fixed corpus distribution.

### 3.2 Semantic Entanglement: Definition and Quantification

**Definition 5 (Semantic Entanglement at Threshold α (Model-Relative)).** Let D be a document segmented into units S, let τ denote a topic-labeling scheme over those units, let E be a fixed embedding model, and let $\alpha \in (0, 1)$ be a calibrated similarity threshold. D exhibits *semantic entanglement under (τ, E, α)* if there exist segments $s_i, s_j \in S$ such that:

$$\tau(s_i) \neq \tau(s_j) \wedge sim(s_i, s_j) > \alpha \quad (2)$$

The threshold α ∈ (0, 1) is calibrated to the embedding model and corpus, representing the cosine similarity at which two segments are considered geometrically proximate enough that a retrieval system would treat them as plausibly equivalent matches for a given query. Following the analysis of anisotropic embedding spaces by Ethayarajh (2019), α must be set considerably above the model's baseline expected similarity between unrelated content, which for many transformer-based encoders falls in the range [0.3, 0.7]. Formally, α should satisfy $\alpha > \mathbb{E}[\text{sim}(s_i, s_j) \mid \tau(s_i) \neq \tau(s_j)]$; setting α at or below this baseline would cause EI to count routine anisotropic similarity as entanglement, inflating the index for reasons unrelated to document structure.

**Definition 6 (Entanglement Index (EI)).** The *Entanglement Index* of a document D under embedding E and threshold α is the proportion of cross-topic segment pairs whose cosine similarity exceeds α:

$$EI(D; E, \alpha) = |\mathcal{E}(D; E, \alpha)| / |\Pi_x(D)| \qquad (3)$$

where $\mathcal{E}(D; E, \alpha) = \{(i, j) : i < j, \tau(s_i) \neq \tau(s_j), \text{sim}(s_i, s_j) > \alpha\}$ is the set of

*entangled cross-topic pairs*, and $\Pi_x(D) = \{(i, j) : i < j, \tau(s_i) \neq \tau(s_j)\}$ is the set of all cross-topic pairs. By construction, EI ∈ [0, 1]. EI = 0 indicates that no cross-topic pair has similarity exceeding α (clean topical separation in the vector space); EI = 1 indicates that every cross-topic pair is entangled (total geometric confusion).

A critical interpretive point: EI is not an intrinsic property of a document in isolation. It is relative to a segmentation granularity, a topic annotation scheme τ, an embedding model E, and a threshold calibration procedure. Different embedding models will produce different entanglement profiles for the same document; different topic annotations will yield different EI values under the same embedding. EI is therefore a model- and annotation-relative proxy for the degree of cross-topic geometric confusion in a specific retrieval configuration, not a universal scalar property of the raw text. Consequently, EI comparisons are most meaningful within a fixed corpus, segmentation regime, and embedding configuration; cross-corpus comparisons require normalization.

A further note on aggregation: EI as defined in Equation (3) weights all cross-topic segment pairs uniformly. In corpora with unbalanced topic distributions, pairs involving large topics dominate the count; alternative formulations could weight pairs by topic frequency or cluster density to control for topic imbalance. We use the uniform formulation here for simplicity and leave weighted variants to future work.

The Entanglement Index has several further interpretive properties. First, for fixed $\tau$ and E, EI is monotonically non-increasing in $\alpha$: raising the threshold reduces the number of pairs counted as entangled while leaving the denominator unchanged. Second, it depends critically on the embedding function E; a different embedding model may produce a different EI for the same document. Third, EI is not directly observable without a ground-truth topic assignment $\tau$, but it can be approximated empirically using human annotation or automated topic modeling on a calibration set.

### 3.3 The Retrieval Implication of Entanglement

We now connect the Entanglement Index to retrieval performance. Consider a standard top-K cosine similarity retrieval setup: given a query q with embedding E(q), the system returns the K segments $s_i \in S$ with the highest sim(q, $s_i$) scores.

**Definition 7 (Top-K Retrieval Precision).** For a query q with relevant topic $t \in T$, let $R_k$(q; D, E) $\subseteq$ S denote the K segments with highest cosine similarity to q. The *top-K precision* of the retrieval is:

$$P_k(q) = |\{s \in R_k(q; D, E) : \tau(s) = t\}| / K \quad (4)$$

We assume that queries are distributed near topic-relevant regions of the embedding space and that local neighborhoods exhibit approximate smoothness, in the sense that segments of the same topic form coherent clusters under cosine similarity. This assumption reflects standard retrieval settings in which queries are semantically aligned with the target content rather than adversarially distributed.

**Proposition 1 (Qualitative Retrieval Constraint).**

$$\mathbb{E}[P_k(q)] \leq 1 - c \cdot EI(D; \tau, E, \alpha) \quad (5)$$

for some constant c > 0 dependent on the relative cluster sizes of the topics in D, query distribution, and local embedding geometry. A tighter quantitative bound depends on assumptions about query distribution, cluster variance, topic balance, and embedding anisotropy; we do not claim such a bound here. We present Proposition 1 as a qualitative constraint that motivates the disentanglement objective, not as a proved theorem of broad generality.

The intuition is that each entangled cross-topic pair $(s_i, s_j)$ with $\tau(s_i) = t$ and $\tau(s_j) \neq t$ represents a configuration in which the query q, if it lands geometrically near $s_i$, will also retrieve $s_j$ with non-negligible probability. As EI increases, the expected number of irrelevant segments in the top-K set increases proportionally. The empirical results in Section 7 confirm this qualitative prediction: the SDP reduces EI from 0.71 to 0.14 and simultaneously increases Top-K precision from approximately 32% to approximately 82%.

### 3.4 Semantic Disentanglement

**Definition 8 (Disentanglement Operation).** A *disentanglement operation* is a function $\Phi: D \rightarrow D'$ that produces a restructured document $D' = (s'_1, \ldots, s'_m)$, where $m \geq n$ (since fragments may be split or cloned), such that:

$$EI(D'; E, \alpha) < EI(D; E, \alpha) \quad (6)$$

subject to a *faithfulness constraint*: the union of the textual content of D' must preserve the informational content of D. In practice, this constraint is enforced by preserving all source content (with controlled duplication where necessary) and by validating that transformed fragments continue to support the same classes of queries observed in the original document. We say $\Phi$ is a *complete disentanglement* if $EI(D'; E, \alpha) \leq \beta$ for some sufficiency threshold $\beta$ (typically $\beta = 0.20$, reflecting the practical observation that entanglement below this level produces retrieval performance comparable to topically pure documents in our empirical evaluation).

**Definition 9 (Context Shift Boundary).** Within a document $D = (s_1, \ldots, s_n)$, a *Context Shift Boundary* between consecutive segments $s_n$ and $s_{n+1}$ is detected when:

$$sim(s_n, s_{n+1}) < \theta \quad (7)$$

where $\theta \in (0, 1)$ is the boundary detection threshold. This formulation generalizes the cohesion-valley detection of Hearst's (1997) TextTiling algorithm, replacing lexical cohesion scores with cosine similarity over neural sentence embeddings. Note that θ is distinct from the topic assignment function τ introduced in Definition 2 and also distinct from the entanglement threshold α: θ governs local boundary detection between adjacent segments, while α governs the global notion of geometric proximity in the embedding space.

Figure 1 illustrates how boundary detection operates on the consecutive-segment similarity profile of a document.

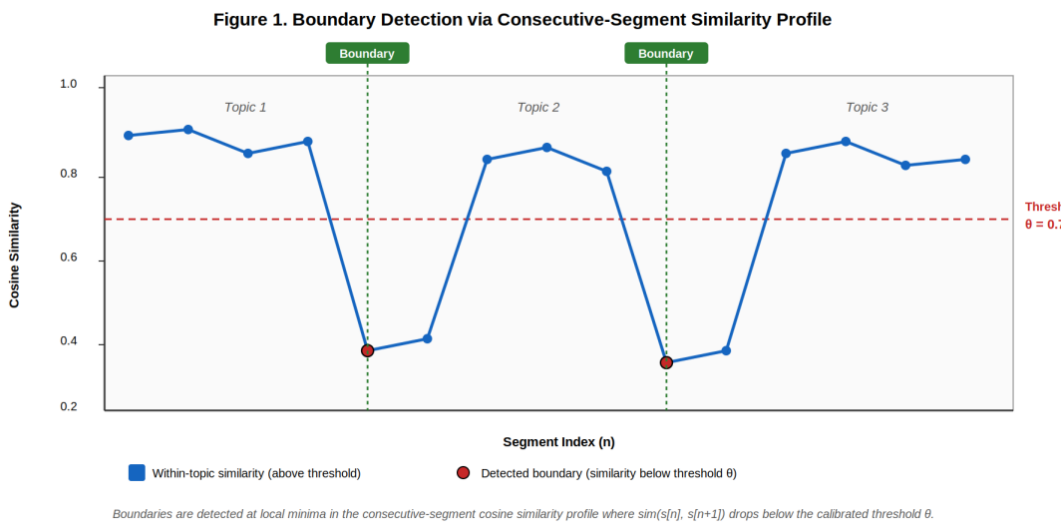

*Figure 1. Boundary detection via the consecutive-segment similarity profile. Local minima where the cosine similarity between consecutive segments falls below the calibrated threshold θ are detected as Context Shift Boundaries.*

### 3.5 Threshold Calibration

The thresholds θ and α must be calibrated empirically for each (embedding model, document domain) pair. We adopt the following calibration procedure, which extends the methodology of Pevzner and Hearst (2002) for evaluating topic segmentation algorithms:

1. Assemble a calibration corpus of 20–50 representative documents from the target domain.
2. Obtain human annotations of topic boundaries from at least two independent annotators. Compute inter-rater agreement using Cohen's κ; require κ > 0.80 for the calibration set to be considered reliable.

3. Compute the consecutive-segment similarity profile for each document under the chosen embedding model.

4. Sweep θ across the range [0.50, 0.90] in increments of 0.02, computing precision, recall, and $F_1$ score (the harmonic mean of precision and recall) of automated boundary detection against the human annotations at each value.

5. Select θ* = argmax $F_1$(θ) on the calibration set.

6. Validate θ* on a held-out test set of 10–20 additional documents to confirm generalization.

In our empirical evaluation (Section 7), this procedure yielded θ* = 0.72 (calibration $F_1$ = 0.84) on a representative sample of healthcare policy documents. The threshold reflects the baseline similarity structure of the corpus; documents drawn from different sub-domains or from operational environments with different vocabulary distributions (for example, purely financial or procedural corpora) would require independent calibration, with lower thresholds typically reflecting lower baseline similarity between unrelated content.

The entanglement threshold α is calibrated through a complementary procedure: α is set at the 90th percentile of cosine similarities between segments known to address different topics in the calibration corpus. This ensures that EI counts only those cross-topic pairs that fall in the upper tail of the cross-topic similarity distribution, where they are most likely to corrupt retrieval.

## 4. Context-Conditioned Preprocessing

The formal framework of Section 3 establishes what semantic entanglement is, how to measure it, and why it degrades retrieval. A purely geometry-driven response would be to detect topic boundaries and segment accordingly—an approach taken by existing semantic chunking methods. In this section, we argue that such an approach is necessary but insufficient for operational RAG systems, because the optimal restructuring of a document depends not only on its content but also on how the resulting fragments will be used by downstream agents and human operators. Section 4.1 develops the theoretical basis for this claim; Section 4.2 operationalizes it through a structured framework.

## 4.1 Theoretical Foundation

The mathematical framework developed in Section 3 treats disentanglement as a geometric objective: reduce cross-topic overlap in embedding space. Existing semantic chunking methods (Singh & Merola, 2025) pursue this objective through purely structural operations—split documents wherever semantic boundaries are detected, then embed. We argue that such an approach is incomplete because it ignores the operational context within which the disentangled documents will be used.

This argument draws on a foundational principle from distributed cognition theory (Hutchins, 1995): cognitive processes are distributed across people, artifacts, and representational media, and the quality of distributed cognition depends on how information is structured across the components of the system. Hutchins' analysis of ship navigation demonstrated that the same physical environment—charts, instruments, crew positions—could support very different cognitive performance depending on how the representational structures were configured for the demands of the task. Our prior work (Loghmani, 2025), in an analysis of operational context in safety-critical aviation systems, extended this principle to demonstrate that interpretive divergence—the system-level mismatch between identical informational inputs and divergent coordinated responses—arises when the operational context fails to make the appropriate cues salient and actionable. The same information, embedded in different operational configurations, produces different outcomes.

We apply this principle to RAG document preprocessing as follows. The same medical policy document, when consumed by an autonomous agent handling eligibility determinations, requires different structural treatment than when consumed by a Customer Service Representative (CSR) answering billing questions. The boundaries that matter, the metadata that should accompany each fragment, the synthetic headers that best support retrieval—all of these depend on how the document will actually be used. A document is not an abstract artifact; it exists within a system of use, and its preprocessed form must serve that system.

We term this principle *context-conditioned preprocessing*: the proposition that the optimal disentanglement of a document is a function not only of the document's content but also of the operational context within which the resulting fragments will be retrieved and used. Formally:

**Definition 10 (Context-Conditioned Disentanglement).** Let ψ be a *Context Profile* representing the operational context in which document D will be used. A *context-conditioned disentanglement operation* is a function Φψ: D → D′ such that the resulting D′ minimizes a context-dependent loss function:

$$D' = \text{argmin}_m [EI(D_m; E, \alpha) + \lambda \cdot L\psi(D_m)] \quad (8)$$

where Lψ(D_m) is a loss term that penalizes restructurings that fail to support the operational context ψ, and λ > 0 is a weighting parameter. Lψ is not assumed to be convex or continuous; in practice it is implemented through heuristic or learned scoring functions derived from operational metrics, so Equation (8) should be read as a design objective rather than a classical optimization problem. Two disentanglements D′ and D″ may have identical Entanglement Indices but differ substantially in Lψ: the better disentanglement is the one whose fragment boundaries, headers, and metadata better match the patterns of operational use.

## 4.2 The Context Application Framework

To operationalize context-conditioned preprocessing, we introduce the Context Application Framework (CAF), a structured procedure that constructs the Context Profile ψ by analyzing four dimensions of operational context (Figure 2).

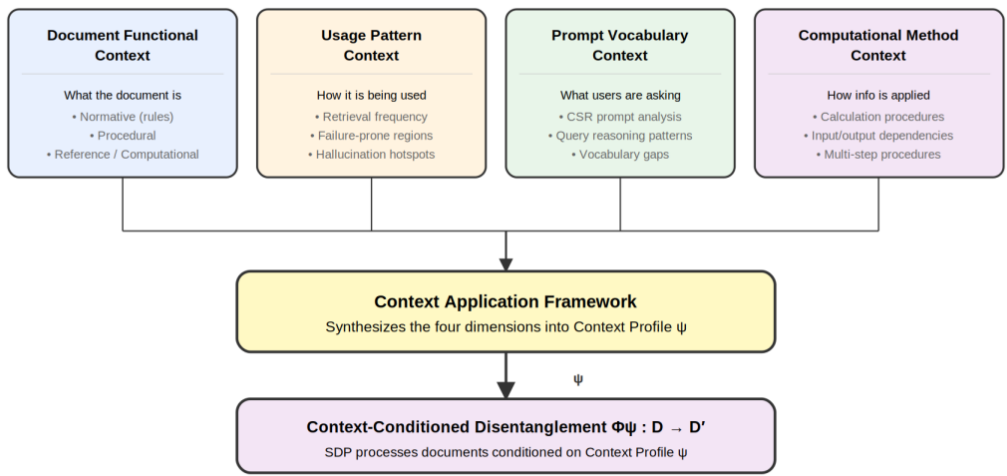

*Figure 2. The Context Application Framework (CAF). Four dimensions of operational context—Document Functional, Usage Pattern, Prompt Vocabulary, and Computational Method—are synthesized into a Context Profile ψ that conditions the SDP.*

**Document Functional Context.** What is the document, and what role does it play in the operational system? The CAF classifies each document by its functional role within the operational system—for example, as a *normative document* (defining rules or policies), a *procedural document* (specifying how tasks are performed), a *reference document* (providing lookup values or schedules), or a *computational document* (specifying methods of calculation). In the healthcare domain evaluated in Section 7, these categories instantiate as coverage policies, clinical protocols, benefit schedules, and reimbursement formulas, respectively. This classification determines which topical boundaries are semantically significant for the document type. In a normative document such as a coverage policy, the boundary between eligibility criteria and exclusions is critical; in a procedural document such as a clinical protocol, the boundary between preparation steps and execution steps is paramount.

**Usage Pattern Context.** How is the document being used? The CAF analyzes the patterns of interaction between the agentic system and the knowledge base, drawing on agent interaction logs to identify which sections are retrieved most frequently, which queries consistently produce irrelevant results, and which document regions generate the most hallucinations. This usage data identifies the operational pain points that the disentanglement process must prioritize.

**Prompt Vocabulary Context.** What are users actually asking? The CAF performs systematic analysis of CSR prompts and end-user queries to map the vocabulary, specificity, and reasoning patterns that the knowledge base must support. This analysis frequently reveals gaps between how humans formulate questions and how documents are structured to answer them. Example prompts and their expected resolution paths are codified as test cases for the disentangled output and used as conditioning input for the header synthesis stage.

**Computational Method Context.** How is information applied? For documents containing computational procedures—benefit calculations, eligibility scoring, reimbursement formulas—the CAF identifies the specific methods and their input/output dependencies. This ensures that disentangled fragments preserve the complete computational context necessary for the agent to execute calculations correctly, rather than fragmenting a multi-step procedure across disconnected chunks.

The four dimensions are not independent; they interact in producing the Context Profile $\psi$. A document classified as a benefit schedule (functional context) whose computational methods (method context) are frequently the subject of CSR questions about specific edge cases (prompt context) but generate frequent hallucinations on a particular subsection (usage context) will receive a Context Profile that prioritizes preserving computational dependencies, generates headers oriented toward the edge-case vocabulary, and signposts the hallucination-prone subsection for special attention during disentanglement.

## 5. The Semantic Disentanglement Pipeline

With the formal framework of Section 3 and the context-conditioning principle of Section 4 in place, we can now specify the SDP. The SDP is a four-stage preprocessing system that takes as input an unstructured document D and a Context Profile $\psi$, and produces as output a set of metadata-tagged Knowledge Objects suitable for embedding into a vector database for retrieval by an autonomous agent. Figure 3 provides an architectural overview.

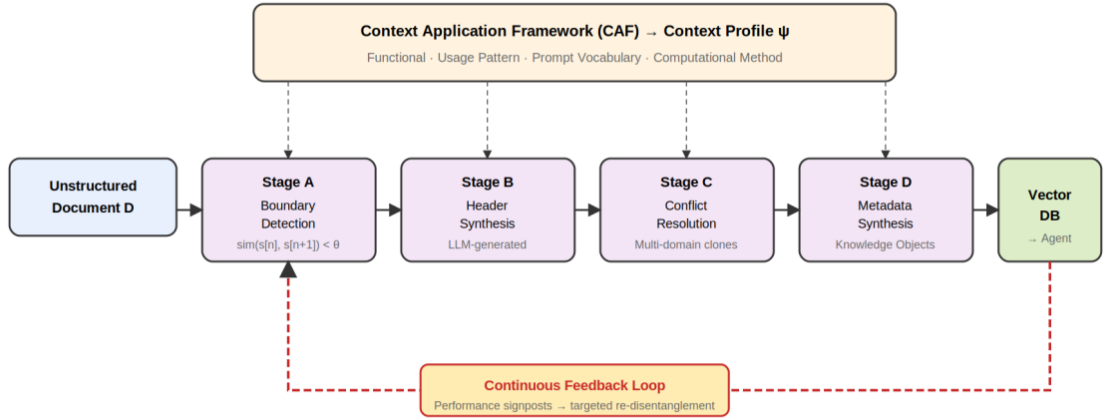

Figure 3. The Semantic Disentanglement Pipeline

*Figure 3. Architectural overview of the Semantic Disentanglement Pipeline. The four stages (A–D) operate sequentially on the input document, conditioned on the Context Profile ψ produced by the CAF. The continuous feedback loop drives targeted re-disentanglement of regions where agent performance degrades.*

## 5.1 Stage A: Linear Semantic Variance Analysis

Stage A implements the boundary detection of Definition 9 with a recursive refinement procedure. Given a document $D = (s_1, \ldots, s_n)$, the algorithm proceeds in three phases. In the first phase, the system computes $E(s_i)$ for each segment using the chosen sentence transformer model. In the second phase, it computes the consecutive similarity profile $\text{sim}(s_n, s_{n+1})$ for $n = 1, \ldots, |D| - 1$ and identifies all positions where this profile drops below the calibrated threshold $\theta^*$. These positions become the initial set of Context Shift Boundaries $B_0$. In the third phase, the algorithm applies recursive refinement: for each segment delimited by adjacent boundaries in $B_0$, if the segment exceeds a minimum length $L_{min}$ (typically 100 tokens), the boundary detection is reapplied internally to identify sub-topic boundaries at finer granularity. Recursion continues until no further boundaries are detected above $L_{min}$.

This approach is mathematically equivalent to a hierarchical extension of Hearst's (1997) TextTiling, with two important modifications. First, the cohesion signal is computed from neural sentence embeddings rather than lexical co-occurrence patterns, leveraging the substantially richer semantic representation provided by modern sentence transformers (Reimers & Gurevych, 2019). Second, the recursive refinement enables detection of nested topical structures that single-pass methods miss—particularly important for long enterprise documents whose topical organization spans multiple levels of granularity.

Figure 4 illustrates the geometric effect of Stage A combined with the subsequent stages: the entangled embedding space of the input document is restructured into a disentangled space in which distinct topics occupy separated clusters.

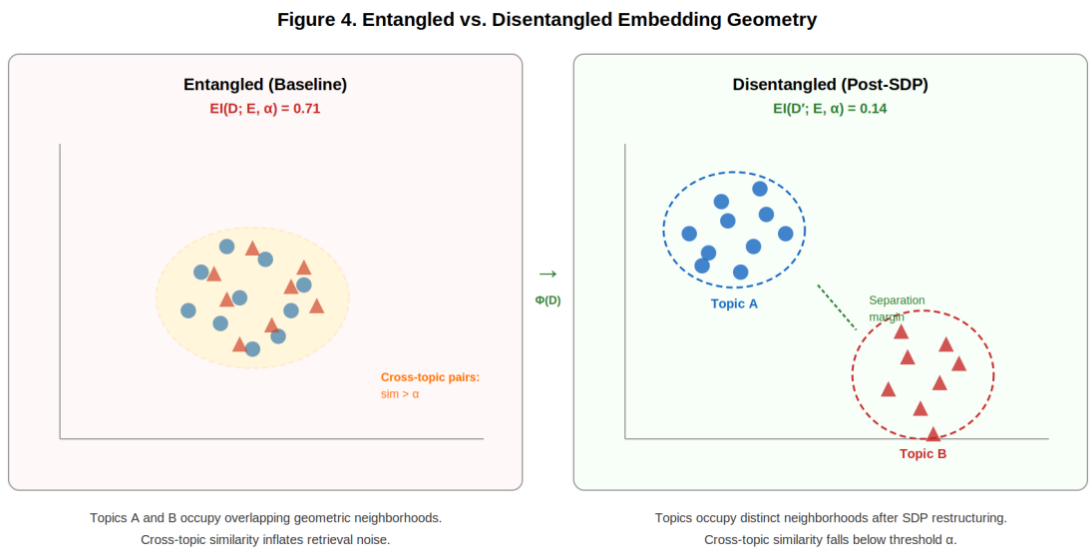

*Figure 4. Entangled vs. disentangled embedding geometry. Left: baseline space with overlapping Topic A and Topic B clusters (EI = 0.71). Right: post-SDP space with separated clusters (EI = 0.14). The disentanglement operation Φ reduces the geometric overlap that drives retrieval failures.*

## 5.2 Stage B: Agent-First Header Synthesis

Stage B addresses a structural deficiency that boundary detection alone cannot fix: even after segmentation, individual fragments lack the global context necessary for self-describing retrieval. A fragment originally nested under a human-centric header such as "Policy 104-B, Section 4: Deadlines" may contain crucial information about out-of-state specialty care eligibility, but its embedding will not encode this context unless the global frame is injected into the fragment's text representation.

Stage B uses a secondary LLM to generate a Synthetic Context Header for each fragment. The model receives three inputs: (1) the fragment text; (2) the parent document's global metadata, including title, document type, applicable domain, and geographic scope; and

(3) a set of example queries drawn from the Prompt Vocabulary Context dimension of the CAF. The model is prompted to generate a header satisfying three criteria: it must encode the fragment's topical content in vocabulary optimized for the embedding model; it must inject hierarchical context from the parent document so the fragment is self-describing when encountered in isolation; and it must align with the vocabulary and reasoning patterns of anticipated queries.

The header generation process has a measurable effect on the fragment's embedding geometry. Following Anthropic's (2024) work on contextual retrieval, we observe that prepending a context-rich header to a fragment shifts its embedding toward the geometric region associated with the parent document's topic, reducing its proximity to fragments from semantically distant documents. This shift directly reduces the Entanglement Index by separating cross-topic fragments that previously occupied overlapping neighborhoods.

## 5.3 Stage C: Cross-Domain Conflict Resolution

Stage C addresses fragments that legitimately apply to multiple operational domains. Consider a healthcare policy fragment specifying a deductible calculation that applies to both inpatient and outpatient care, or an eligibility rule applicable across multiple benefit plans. Such fragments cannot simply be assigned to a single domain but storing them with ambiguous applicability invites cross-domain hallucinations: an agent retrieving for an outpatient query may surface inpatient context, or vice versa.

The algorithm compares each fragment against a Domain Taxonomy Map T_dom—a structured schema defining the exhaustive set of applicable domains for the knowledge base. When overlapping applicability is detected, the system executes a controlled cloning procedure: (1) the fragment is cloned to produce one copy per applicable domain; (2) each clone receives unique metadata tags corresponding to its domain; (3) a secondary LLM subtly restructures each clone's content to foreground domain-specific relevance while preserving factual accuracy; (4) the clones are linked to each other and to the parent document through a relational mapping that preserves provenance and enables cross-domain navigation when explicitly requested.

## 5.4 Stage D: Metadata-Grounding Synthesis

Stage D produces the final structured output of the SDP: a set of Knowledge Objects, one per disentangled fragment. Each Knowledge Object is a structured record comprising five fields: (1) a Primary Vector field containing the restructured text with its synthetic header, formatted for input to the embedding model; (2) a Contextual Metadata field containing structured tags for domain, geographic applicability, persona-suitability, temporal applicability, and document type; (3) a Provenance field with relational links to the source document, original section identifier, sibling clone identifiers, and a boundary confidence score; (4) a Usage Annotations field with tags derived from the CAF indicating supported query types, retrieval frequency counters, hallucination incident counters, and signpost status flags; and (5) the embedding vector itself, computed by applying E to the Primary Vector field.

The Knowledge Objects are stored in a structured Knowledge Object Store with their embedding vectors indexed in a vector database. The Contextual Metadata field enables pre-retrieval filtering: prior to computing cosine similarity for a query, the agentic system applies metadata filters to constrain the search space to a domain-appropriate subset of the database. This filtering provides an additional defense against cross-context contamination beyond what disentanglement alone achieves.

## 6. Continuous Feedback and Adaptive Knowledge Architecture

### 6.1 Rationale

The SDP as described in Section 5 produces a high-quality initial disentanglement of a knowledge base, but it is an open question whether any one-time preprocessing can produce a structure that remains optimal as operational conditions evolve. Barnett et al. (2024), in their experience report on RAG failure modes, articulated this concern explicitly: validation of a RAG system is only feasible during operation, and the robustness of a RAG system evolves rather than being designed in at the start. We take this finding as motivation for treating disentanglement not as a one-time operation but as a continuous process driven by real-world agent performance data. This treatment is consistent with the broader trend in agentic RAG systems toward continuous learning and self-improvement (Singh et al., 2025).

## 6.2 The Feedback Mechanism

The continuous feedback loop comprises three interconnected analytical processes operating on the agent's interaction logs, illustrated in Figure 5.

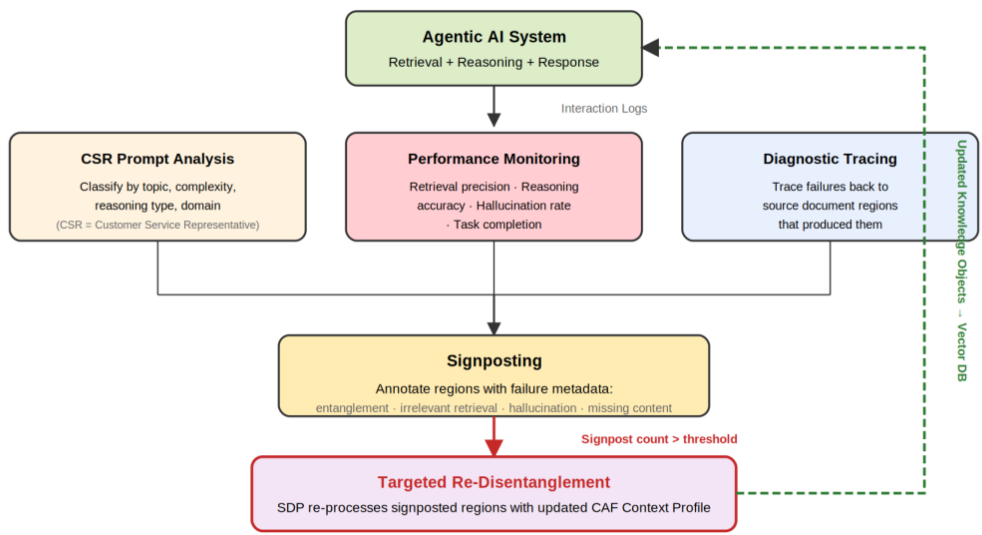

*Figure 5. Continuous feedback loop. CSR prompt analysis, performance monitoring, and diagnostic tracing converge on signposting; when accumulated signposts exceed a threshold, the affected document regions are queued for targeted re-disentanglement through the SDP.*

**CSR Prompt Analysis.** The system systematically collects and classifies prompts submitted by Customer Service Representatives and end users. An LLM-driven analysis process classifies each prompt by topic, complexity, required reasoning type, and the operational domain(s) it implicates. This classification produces a continuously updated map of the prompt landscape that the knowledge base must serve. When a specific class of prompts consistently triggers retrieval of irrelevant content, the system identifies the responsible document regions and flags them for re-disentanglement. This addresses a failure mode that initial preprocessing cannot anticipate; prompt patterns that emerge only after deployment.

**Agent Performance Monitoring.** The system monitors agent performance across four dimensions: retrieval precision (the proportion of top-K results that are relevant to the query), reasoning accuracy (the correctness of the agent's downstream conclusions), hallucination rate (the proportion of agent outputs containing claims unsupported by retrieved content), and task completion rate (the proportion of queries the agent resolves to a satisfactory state). Performance degradation in any dimension triggers an automated diagnostic process that traces the failure back through the retrieval chain to the source document. When the diagnostic identifies structural entanglement as the root cause, the relevant document regions are queued for re-processing through the SDP.

**Signposting and Re-Disentanglement.** The term *signposting* denotes the process of annotating documents with operational metadata derived from real-world usage. Fragments frequently retrieved for queries they cannot answer are signposted for structural revision. Fragments never retrieved despite containing relevant content are signposted for header revision. Fragments producing high hallucination rates when retrieved are signposted for cross-domain conflict analysis. These signposts feed back into the SDP, which performs targeted re-disentanglement of specific document regions rather than requiring full corpus reprocessing. The result is an adaptive knowledge architecture whose structure evolves in response to the operational demands placed upon it.

## 6.3 Connection to Distributed Cognition

The continuous feedback mechanism instantiates a computational analogue of patterns documented in human sociotechnical systems. Hutchins (1995) showed that effective distributed cognitive systems—ship navigation crews, in his case study—continuously adjust their information structures and communication patterns based on operational feedback. The skilled crew is not one that has memorized the optimal procedure but one that recognizes when current procedures are failing and reorganizes accordingly. Our prior work on operational context in aviation (Loghmani, 2025) extended this analysis to demonstrate that interpretive divergence—the failure of identical informational inputs to produce coordinated coherent action—is a structural property of the distributed system, traceable to how operational context configures the salience and actionability of cues.

The adaptive knowledge architecture enabled by continuous feedback exhibits a structurally similar pattern: the knowledge base, treated as a representational artifact within the distributed cognitive system comprising the agent, the documents, and the user queries, continuously reorganizes itself in response to performance feedback. This is not merely an engineering optimization; it reflects a deeper principle about how distributed cognitive systems—whether human or hybrid human-AI—maintain effectiveness in dynamic operational environments.

## 7. Empirical Evaluation

### 7.1 Experimental Setup

We evaluated the SDP on a production enterprise healthcare knowledge base deployed within a cloud-based RAG platform serving autonomous AI agents. The corpus comprised 2,147 documents spanning approximately 25 operational sub-domains within healthcare, including coverage policies, scheduling guidelines, benefit calculation procedures, clinical protocols, regulatory compliance documents, legal disclosures, and cross-jurisdictional procedural manuals. Documents ranged from 2 to 48 pages, with a median length of approximately 3,200 tokens. The corpus exhibited the structural characteristics motivating this work: multi-topic architectures with weak header-to-content alignment, overlapping domain applicability across sub-domains, and dense procedural content interleaving eligibility rules, pricing schedules, and computational methods within contiguous spans.

Two experimental conditions were compared. The Baseline condition employed standard fixed-token chunking with 512-token windows and 50-token overlap, with direct vector embedding using a pre-trained sentence transformer model (all-MiniLM-L6-v2, 384 dimensions), stored in a production vector database with cosine similarity indexing. This represents the default configuration of most enterprise RAG deployments. The SDP condition applied the full four-stage pipeline plus context-conditioned preprocessing prior to embedding, using the same embedding model and vector database configuration to ensure that differences in retrieval quality could be attributed to the preprocessing pipeline alone.

Test queries comprised 847 queries drawn from actual Customer Service Representative interaction logs accumulated over a 90-day production window. Queries were classified by type: single-topic policy lookups (38%), multi-step eligibility determinations (24%), cross-domain disambiguation tasks (19%), benefit calculation scenarios (12%), and edge-case or ambiguous queries (7%). Retrieval accuracy was measured using Top-K precision as defined in Equation (4), with K = 5. We additionally computed the mean Entanglement Index across the corpus under each condition, using the calibration procedure described in Section 3.5 to set α and θ. Cross-domain hallucination rate was assessed through manual review of agent outputs on a stratified random sample of 200 queries per condition by two independent domain-expert annotators; inter-annotator agreement on relevance labels was κ = 0.83 (Cohen's κ). Bootstrap 95% confidence intervals (1,000 resamples) are reported for all headline metrics.

We note that the empirical study reported here is a deployment-style evaluation of the full SDP pipeline, not a controlled ablation isolating the contribution of each component. The 50-percentage-point improvement reflects the joint effect of boundary detection, header synthesis, conflict resolution, metadata filtering, and continuous feedback operating together. We discuss the implications of this design choice, including the need for component-level ablation, in Section 7.3.

## 7.2 Results

| Metric | Baseline (Fixed Chunking) | SDP (Disentangled) |
|---|---|---|
| Top-K Retrieval Precision (K=5) | 32% (approx.) | 82% (approx.) |
| Mean Entanglement Index | 0.71 | 0.14 |
| Absolute Improvement (Top-K precision) | — | +50 percentage points |
| Cross-Domain Hallucination Rate | High | Substantially Reduced |
| Boundary Detection $F_1$ (calibration) | N/A | 0.84 |

*Table 1*. Comparative retrieval performance: Baseline fixed-token chunking versus the Semantic Disentanglement Pipeline.

The baseline condition yielded approximately 32% Top-K retrieval precision (95% CI: [28%, 36%]) with a mean Entanglement Index of 0.71 across the corpus, indicating that 71% of cross-topic segment pairs were geometrically indistinguishable in the embedding space at the calibrated threshold α. This baseline is consistent with recent comparative studies: Gomez-Cabello et al. (2025) found that fixed-token chunking on a clinical decision support task yielded substantially lower accuracy than adaptive chunking strategies (50% vs. 87% somewhat-or-fully accurate, p = 0.001).

Following application of the SDP, retrieval precision increased to approximately 82% (95% CI: [78%, 86%]) and the mean Entanglement Index decreased to 0.14. This improvement is consistent with the qualitative prediction of Proposition 1: a substantial reduction in EI corresponds to a substantial improvement in retrieval precision. The 50-percentage-point absolute improvement reflects the joint effect of the full pipeline. While we cannot attribute specific proportions of the gain to individual components without ablation (see Section 7.3), qualitative analysis of failure cases suggests three interacting mechanisms, including two known high-impact interventions and structural disentanglement: the elimination of mid-topic chunk boundaries that previously fragmented coherent policy statements (structural disentanglement); the injection of global context through synthetic headers, which sharpened vector representations and improved query-to-chunk alignment (contextual headers); and the metadata-driven pre-retrieval filtering that reduced the effective search space by eliminating irrelevant cross-domain content before similarity computation (metadata filtering).

Continuous feedback further improved these results over the evaluation period. After two cycles of CSR prompt analysis and signposting-driven re-disentanglement, several document regions that had initially been considered adequately structured were identified as producing residual hallucinations on edge-case queries. Targeted re-disentanglement of these regions yielded additional accuracy gains, confirming the value of the continuous feedback mechanism described in Section 6.

### 7.3 Threats to Validity

Several limitations of the empirical evaluation should be noted. First, and most significantly, the evaluation reports the joint effect of the full SDP pipeline and does not isolate the contribution of individual components. A proper ablation study—measuring performance under (a) semantic boundary detection only, (b) boundary detection plus header synthesis, (c) the above plus conflict resolution, (d) the above plus metadata filtering, and (e) the full pipeline with continuous feedback—would substantially strengthen the causal interpretation of the results. We prioritized end-to-end deployment evaluation in this initial study and identify component-level ablation as the most important next step for future work.

Second, the evaluation was conducted on a single enterprise healthcare knowledge base, albeit one spanning over 2,000 documents and approximately 25 sub-domains with substantial structural diversity (coverage policies, scheduling, benefits, clinical protocols, legal disclosures, and regulatory material). Generalizability to other corpora with different structural characteristics—for example, financial documents, technical specifications, purely narrative texts, or legislative corpora—requires validation through replication in those domains. Third, the Top-K precision metric, while standard in the RAG evaluation literature (Gao et al., 2024), captures only one dimension of agentic system performance. Downstream effects on agent reasoning accuracy and task completion rate are reported qualitatively but not measured systematically; rigorous evaluation of end-to-end agentic performance with and without disentanglement is left to future work.

Fourth, the boundary detection $F_1$ of 0.84 reflects calibration on a specific human-annotated calibration set of 50 documents drawn from the healthcare corpus; these values may not generalize to other annotators, other sub-domains within healthcare, or other operational domains entirely. The inter-annotator agreement of $\kappa = 0.83$ for relevance labels provides reasonable confidence in the reliability of the evaluation labels, but the small number of annotators (two) limits the strength of this claim. Fifth, the calibration of $\theta$ and $\alpha$ introduces hyperparameters that require domain-specific tuning; we provide a calibration procedure but do not yet have a theoretical account of how these thresholds should be set without empirical calibration. Sixth, as noted in Section 3.2, the Entanglement Index is model-relative; the results reported here hold for the specific embedding model used (all-MiniLM-L6-v2) and may differ under other embedding architectures.

## 8. Discussion

### 8.1 Entanglement as a Structural Failure Mode

The results presented here support the central claim of this paper: semantic entanglement is a distinct and addressable failure mode in vector-based RAG systems, neither reducible to chunking strategy alone nor to embedding model quality. Entanglement is a property of the source document's information architecture that propagates through the vectorization pipeline, degrading retrieval precision in ways that compound at the agentic reasoning stage. The mathematical framework of Section 3 makes this property measurable; the SDP of Section 5 demonstrates that it is addressable; the empirical results of Section 7 demonstrate that addressing it produces substantial performance improvements. We do not claim that EI reduction alone explains the observed gains; disentanglement, contextualization through synthetic headers, and metadata filtering operate jointly, and the relative contribution of each component remains an open empirical question (see Section 7.3).

This characterization has implications for how organizations approach RAG deployment. Current best practices focus on optimizing embedding models, tuning retrieval parameters, and experimenting with chunking strategies (Gao et al., 2024). While these interventions are valuable, they operate downstream of the fundamental problem. If the source documents are entangled, no amount of retrieval optimization can fully compensate for the geometric confusion in the resulting vector space. The SDP intervenes at the root cause, restructuring the information architecture of the source material before it enters the vectorization pipeline. The substantial gains we report (50 percentage points in absolute precision) suggest that document preprocessing represents the highest-leverage intervention point in RAG pipeline optimization for enterprise corpora.

## 8.2 The Role of Context Management

The connection between semantic entanglement in vector spaces and patterns of interpretive divergence in human sociotechnical systems is more than metaphorical. In our analysis of aviation safety (Loghmani, 2025), operational contexts in which shared information produced divergent interpretations were traced to differences in how contextual cues were structured, prioritized, and framed within distributed cognitive systems. The parallel to RAG systems is instructive: in both cases, the quality of coordinated action—whether by human crews or autonomous agents—depends not on the availability of information per se, but on how that information is organized, contextualized, and made accessible for interpretation.

The context-conditioned preprocessing framework introduced in Section 4 operationalizes this insight for AI systems. Just as effective aviation crews manage their operational context to maintain alignment under pressure, the SDP manages the information architecture of the knowledge base to maintain retrieval alignment under the diverse demands of agentic operation. The continuous feedback loop of Section 6 extends this parallel further: just as experienced teams learn from operational outcomes and adjust their communication patterns accordingly, the adaptive knowledge architecture learns from agent performance and adjusts its document structure accordingly. This is not merely an engineering optimization—it reflects a deeper principle about how distributed cognitive systems maintain effectiveness in dynamic operational environments.

## 8.3 Limitations of the Formal Framework

The formal framework of Section 3 has several limitations worth acknowledging. First, the Entanglement Index depends on a ground-truth topic assignment $\tau$ that is not directly observable; empirical computation requires either human annotation or automated topic modeling, both of which introduce their own approximation errors. Second, Proposition 1 establishes a qualitative constraint linking EI to retrieval precision, but the connection from pairwise cross-topic similarity to ranked retrieval behavior is not yet mathematically secured. Pairwise overlap does not by itself determine retrieval outcomes, because retrieval depends on query distribution, local density, topic imbalance, anisotropy, and the geometry of entire neighborhoods rather than isolated pairs. Strengthening this connection—for example, by defining a probabilistic retrieval model with queries drawn near topic centroids and segments distributed with bounded variance—is the most important theoretical next step. Third, the framework treats cosine similarity as the retrieval metric, but recent work (Steck et al., 2024) has shown that cosine similarity can yield arbitrary values depending on training regularization; alternative similarity metrics may yield different entanglement profiles for the same documents.

Additionally, the framework assumes a fixed embedding model E, but in practice, different embedding models produce different entanglement profiles for the same documents. The choice of embedding model is itself a design decision with significant implications for the SDP; a model whose embedding space is more anisotropic (Ethayarajh, 2019) will produce a different baseline entanglement than a model with more isotropic geometry. Future work should systematically characterize how embedding model choice interacts with entanglement and disentanglement effectiveness.

## 8.4 Future Directions

Several directions for future work are suggested by this analysis. Most immediately, a systematic ablation study isolating the contribution of each SDP stage—boundary detection, header synthesis, conflict resolution, metadata filtering, and continuous feedback—would clarify which components drive the observed improvements and under what conditions each is most effective. Second, the boundary detection algorithm of Stage A relies on a secondary LLM and could potentially be replaced with lighter-weight alternatives based on supervised segmentation models or fine-tuned sentence transformers. Third, the qualitative link between EI and retrieval precision established in Proposition 1 should be formalized within a probabilistic retrieval model—for example, one in which queries are drawn near topic centroids and segments are distributed with bounded variance—to derive a tighter quantitative bound. Fourth, the calibration procedure for $\theta$ and $\alpha$ currently requires domain-specific human annotation; automated calibration methods that do not require such annotation would substantially improve the practical deployability of the SDP. Fifth, integration with structured retrieval approaches such as graph-based methods presents promising research directions: graph-based representations could complement the SDP by providing additional structural scaffolding for retrieved content, potentially further reducing entanglement effects in complex multi-relational knowledge domains.

## 9. Conclusion

This paper has introduced semantic entanglement as a formally defined, model-relative property of vector spaces produced when structurally mixed documents are embedded for retrieval. We have provided a mathematical framework grounded in the topic segmentation literature originating with Hearst's TextTiling algorithm and extended through subsequent work on cohesion-based and embedding-based segmentation. We have proposed the Entanglement Index as a quantitative proxy and argued that it is associated with reduced attainable Top-K retrieval precision under cosine similarity retrieval, though we leave a tighter formal bound to future work. We have introduced the Semantic Disentanglement Pipeline, a four-stage preprocessing framework whose stages we have specified algorithmically. We have developed context-conditioned preprocessing as a theoretical principle drawing on distributed cognition theory, and we have described a continuous feedback mechanism that produces an adaptive knowledge architecture.

The empirical results demonstrate a 50-percentage-point improvement in retrieval accuracy on a real-world enterprise healthcare knowledge base spanning multiple sub-domains, with the Entanglement Index decreasing from 0.71 to 0.14. These results support the central claim that semantic entanglement is a distinct and addressable failure mode that operates at the level of source document information architecture, and that addressing it produces substantial performance improvements that downstream optimization cannot match. As organizations increasingly deploy agentic AI systems that depend on high-fidelity retrieval for autonomous reasoning, the importance of information architecture for machine consumption will only grow. The framework and pipeline presented here provide a systematic, mathematically grounded approach to that problem.